\def\year{2021}\relax
\documentclass[letterpaper]{article} 
\usepackage{aaai21}  
\usepackage{times}  
\usepackage{helvet} 
\usepackage{courier}  
\usepackage[hyphens]{url}  
\usepackage{graphicx} 
\def\UrlFont{\rm}  
\usepackage{natbib}  
\usepackage{graphicx}  

\usepackage{algorithmicx}
\usepackage{amsmath}
\usepackage{amsfonts}
\usepackage{amssymb}
\usepackage{amsthm}
\usepackage{mathrsfs}
\usepackage{algorithm}
\usepackage[noend]{algpseudocode}
\usepackage[switch]{lineno}
\title{Bayesian Distributional Policy Gradients}
\author {
       Luchen Li,
     \textsuperscript{\rm 1}
        A. Aldo Faisal \textsuperscript{\rm 1,2,3}\\
}
\affiliations {
    \textsuperscript{\rm 1} Brain \& Behaviour Lab, Dept. of Computing, Imperial College London, UK\\
    \textsuperscript{\rm 2}  Brain \& Behaviour Lab, Dept. of Bioengineering, Imperial College London, UK\\
    \textsuperscript{\rm 3} UKRI Centre in AI for Healthcare,
    Imperial College London, UK\\
    $\{$l.li17, aldo.faisal$\}$@imperial.ac.uk
}

\begin{document}
\maketitle
\begin{abstract}
Distributional Reinforcement Learning (RL) maintains the entire probability distribution of the reward-to-go, i.e. the return, providing more learning signals that account for the uncertainty associated with policy performance, which may be beneficial for trading off exploration and exploitation and policy learning in general. Previous works in distributional RL focused mainly on computing the state-action-return distributions, here we model the state-return distributions. This enables us to translate successful conventional RL algorithms that are based on state values into distributional RL. We formulate the distributional Bellman operation as an inference-based auto-encoding process that minimises Wasserstein metrics between target/model return distributions. The proposed algorithm, BDPG (Bayesian Distributional Policy Gradients), uses adversarial training in joint-contrastive learning to estimate a variational posterior from the returns. Moreover, we can now interpret the return prediction uncertainty as an information gain, which allows to obtain a new curiosity measure that helps BDPG steer exploration actively and efficiently. We demonstrate in a suite of Atari 2600 games and MuJoCo tasks, including well known hard-exploration challenges, how BDPG learns generally faster and with higher asymptotic performance than reference distributional RL algorithms.
\end{abstract}

\section{Introduction}
In reinforcement learning (RL), the performance of a policy is evaluated by the (discounted) accumulated future rewards, a random variable known as the reward-to-go or the return. Instead of maintaining the expectations of returns as scalar value functions, distributional RL estimates \textit{return distributions}. Keeping track of the uncertainties around returns has initially been leveraged as a means to raise risk awareness in RL \cite{morimura10, lattimore12}. Recently, a line of research pioneered by \cite{bellemare17} applied the distributional Bellman operator for control purposes. Distributional RL is shown to outperform previous successful deep RL methods in Atari-57 when combined with other avant-garde developments in RL \cite{hessel18, dabney18}.

The critical hurdle in distributional RL is to minimise a Wasserstein distance between the distributions of a return and its Bellman target, under which the Bellman operation is a contraction mapping \cite{bellemare17}. A differentiable Wasserstein distance estimator can be obtained in its dual form with constrained Kantorovich potentials \cite{gulrajani17, arjovsky17}, or approximated by restricting the search for couplings to a set of smoothed joint probabilities with entropic regularisations \cite{cuturi13, montavon16, genevay16, luise18}.
Alternatively, a Bayesian inference perspective
redirects the search space to a set of probabilistic encoders that map data in the input space to codes in a latent space \cite{bousquet17, tolstikhin18, ambrogioni18}. Bayesian approaches rely on inference to bypass rigid and sub-optimal distributions that are usually entailed otherwise, while retaining differentiability and tractability. Moreover, predictions based on inference, the expectation across a latent space, are more robust to unseen data \cite{blundell15} and thus able to generalise better.

In contrast to previous distributional RL work that focuses on state-action-return distributions, here we investigate state-return distributions and prove that its Bellman operator is also a contraction in Wasserstein metrics. This  opens up the possibility of converting  state-value algorithms into distributional RL settings. We then formulate the distributional Bellman operation as an inference-based auto-encoding process that minimises Wasserstein metrics between continuous distributions of the Bellman target and estimated return. A second benefit of our inference model is that the learned posterior enables a curiosity bonus in the form of information gain (IG), which is leveraged as internal reward to boost exploration efficiency. We explicitly calculate the entropy reduction in a latent space corresponding to return probability estimation as a KL divergence. In contrast to  previous work \cite{bellemare16count, sekar20, ball20rpo} in which IG was approximated with ensemble entropy or prediction gains, we obtain analytical results from our variational inference scheme. 

To test our fully Bayesian approach and curiosity-driven exploration mechanism against a distributional RL backdrop, we embed these two innovations into a policy gradient framework. Both innovations  would  also work for value-based and off-policy policy gradients methods where the state-action-return distribution is modelled instead.

We evaluate and compare our method to other distributional RL approaches on the Arcade Learning Environment Atari 2600 games \cite{Bellemare2013ALE}, including some of the best known hard-exploration cases, and on MuJoCo continuous-control tasks \cite{todorov12mujoco}. To conclude we perform ablation experiments, where we investigate our exploration mechanism and the length of bootstrapping in distributional Bellman backup.

Our key contributions in this work are two-fold:
we derive first, a fully inference-based generative approach to distributional Bellman operations; and second, a novel curiosity-driven exploration mechanism formulated as posterior information gains attributed to return prediction uncertainty.

\section{Preliminaries}

\subsection{Wasserstein Variational Inference}
In this subsection, we discuss how Wasserstein metrics in a data space can be estimated in a Bayesian fashion using adversarial training. Notation-wise we use calligraphic letters for spaces, capital letters for random variables and lower-case letters for values. We denote probability distributions and densities with the same notation, discriminated by the argument being capital or  lower-case, respectively.

In optimal transport problems \cite{villani08ot}, divergences between two probability distributions are estimated as the cost required to transport probability mass from one to the other. Consider input spaces $\mathcal{X}\in\mathbb{R}^n$, $\mathcal{Y}\in\mathbb{R}^m$ and a pairwise cost function $c: \mathcal{X}\times\mathcal{Y}\mapsto\mathbb{R}^+$. For two probability measures $\alpha: \mathcal{X}\mapsto\mathscr{P}, ~\beta: \mathcal{Y}\mapsto\mathscr{P}$, an optimal transport divergence is defined as
\begin{equation}
\mathcal{L}_c(\alpha, \beta) := \mathop{\mathrm{inf}}_{\gamma \in\Gamma(\alpha, \beta)} \int_{\mathcal{X}\times\mathcal{Y}} c(x, y)\mathrm{d}\gamma(x, y),
\end{equation}
where $\Gamma(\alpha, \beta)$ is a set of joint distributions or couplings on $\mathcal{X}\times\mathcal{Y}$ with marginals $\alpha$ and $\beta$ respectively. Particularly, when $\mathcal{Y}=\mathcal{X}$ and the cost function is derived from a \textit{metric} over $\mathcal{X}$, $d:\mathcal{X}\times\mathcal{X}\mapsto\mathbb{R}^+$, via $c(x, y) = d^p(x, y), ~p\geq 1$, the $p$-Wasserstein distance on $\mathcal{X}$ is given as
\begin{equation}\label{eq:wass_inf}
W_p(\alpha, \beta) := \mathcal{L}_{d^p}(\alpha, \beta)^{1/p}.
\end{equation}

Now consider a generative process through a latent variable $Z\in\mathcal{Z}\in\mathbb{R}^l$ with a prior $p_Z(Z)$, a decoder $p_\theta(X|Z)$ and an encoder (amortised inference estimator) $q_\phi(Z|X)$, in which the parameters $\phi, ~\theta$ are trained to mimic the data distribution $p_X(X)$ implicitly represented by the i.i.d. training samples. The density corresponding to the model distribution can be expressed as $p_G(x) = \mathbb{E}_{z\sim p_Z}[ p_\theta(x|z)]$. For a deterministic\footnote{For the purpose of generative modelling, the intuition of minimising Wasserstein metrics between target/model distributions (instead of stronger probability density discrepancies such as $f$-divergences) is to still see meaningful gradients when the model manifold and the true distribution’s support have few intersections without introducing noise to the model distribution (by using a directed continuous mapping) that renders reconstructions blurry.} decoder $X=G_\theta(Z)$, $p_G$ can be thought of as the \textit{push-forward} of $p_Z$ through $G_\theta$, i.e. $p_G = G_{\theta\#}p_Z$. Minimising the Wasserstein distance between $p_X(X)$ and $p_G(X)$ is thereby equivalent to finding an optimal transport plan between $p_X(X)$ and $p_Z(Z)$, and matching the aggregated posterior $Q(Z) := \mathbb{E}_{x\sim p_X}\big[q_\phi(Z|x)\big]$ to the prior $p_Z(Z)$
\cite{bousquet17, tolstikhin18, ambrogioni18, rosca18, he19} 
\begin{equation}\label{eq:wae}
W_p^p(p_X, p_G) = \mathop{\mathrm{inf}}_{q_\phi:Q=p_Z} \mathbb{E}_{X\sim p_X}\mathbb{E}_{Z\sim q_\phi} \big[d^p\big(X, G_\theta(Z)\big)\big].
\end{equation}

Marginal matching in $\mathcal{Z}$ is sometimes preferred for generative models, since it alleviates the posterior collapse problem \cite{zhao17InfoVAE, hoffman16elbo} by enabling $Z$ to be more diversely distributed for different $x$'s.
However, when doing so, Eq. \eqref{eq:wae} is no longer 
a proper inference objective, as it enforces neither posterior-contrastive nor joint-contrastive learning. In fact, the encoder needs not to approximate the true posterior $p_\theta(Z|X)$ exactly to satisfy the marginal match. In contrast, our approach maintains a fully Bayesian inference pipeline.

While explicit variational inference requires all probability densities to have analytical expressions, we bypass this by direct density matching through adversarial training \cite{goodfellow14gans}, which requires only that densities can be sampled from for gradient backpropagation, thereby allowing for a degenerate decoder $p_\theta(X|Z)=\delta_{G_\theta(Z)}(X)$.  \\

\noindent
\textbf{Lemma 1.} \cite{donahue17bigan} \textit{Let $p(X, Z)$ and $q(X, Z)$ denote the joint sampling distribution induced by the decoder and encoder respectively, $D_\psi$ a discriminator, and define} $F(\psi, \phi, \theta) := \mathbb{E}_{x, z\sim p} \big[\log D_\psi(x, z)\big]+ \mathbb{E}_{x, z\sim q} \big[\log \big(1 - D_\psi(x, z)\big)\big]$. \textit{For any encoder and decoder, deterministic or stochastic, the optimal discriminator} $D_{\psi^*} = \mathop{\mathrm{argmax}}_{D_\psi} F(\psi, \phi, \theta)$ \textit{is the \\Radon-Nikodym derivative of measure} $p(X, Z)$ \textit{w.r.t.} $p(X, Z) + q(X, Z)$. \textit{The encoder and decoder's objective for an optimal discriminator} $C(\phi, \theta) := F(\psi^*, \phi, \theta)$ \textit{can be\\ written in the Jenson-Shannon divergence} $C(\phi, \theta) = 2 \mathrm{JS}\big(p, q\big) - \log 4$, \textit{in which the global minimum is achieved if and only if} $p(X, Z) = q(X, Z)$.\\

We jointly minimise Wasserstein metrics between model/target distributions in $\mathcal{X}$ and conduct variational inference adversarially. $p_G$ will be shown to be modelling the distribution of random returns, leading to a novel approach to accomplishing distributional Bellman operations.

\subsection{Distributional Reinforcement Learning}
We start by laying out  RL and policy gradients notation, then explain the distributional perspective of RL, as well as previous solutions to it.

\subsubsection{Policy Gradients}
A standard RL task is framed within a Markov decision process (MDP) $\big(\mathcal{S}, \mathcal{A}, \mathscr{R}, P, \gamma\big)$ \cite{puterman94mdp}, where $\mathcal{S}$ and $\mathcal{A}$ denote the state and action spaces respectively, $\mathscr{R}: \mathcal{S}\times\mathcal{A}\mapsto\mathbb{R}^n$ a potentially stochastic reward function, $P: \mathcal{S}\times\mathcal{A}\mapsto\mathscr{P}(\mathcal{S})$ a transition probability density function, and $\gamma\in (0, 1)$ a temporal discount factor. An RL agent has a policy that maps  states to a probability distribution over actions $\pi : \mathcal{S}\mapsto\mathscr{P}(\mathcal{A})$.

The return $G^\pi$ under the policy $\pi$ is a random variable that represents the sum of discounted future rewards and the \textit{state}-dependent return is
$G^\pi(s) := \sum_{t=0}^\infty \gamma^t r_t, s_0 = s$.

A state value function is defined as the expected return $V^\pi(s):=\mathbb{E}[G^\pi(s)]$,
a state-action value function the expected state-action return $Q^\pi(s, a) :=\mathbb{E}[G^\pi(s, a)]$.
The Bellman operator $\mathcal{T}^\pi$ \cite{Bellman1957} is defined as
\begin{equation}\label{eq:bellmanop_convention}
\mathcal{T}^\pi V^\pi(s) := \mathbb{E}_{\pi, \mathscr{R}, P} \big[r + \gamma V^\pi(s')\big],
\end{equation}
\begin{equation}\label{eq:bellmanop_convention_q}
\mathcal{T}^\pi Q^\pi(s, a) := \mathbb{E}_{\mathscr{R}, P, \pi} \big[r + \gamma Q^\pi(s', a')\big].
\end{equation}

Policy gradient methods \cite{62Sutton1999PGM} 
optimise a \\parameterised policy $\pi$ by directly ascending the gradient of\\ a policy performance objective such as $\mathbb{E}_{s\sim d^{\pi}, a\sim\pi(\cdot|s)} \big[\log \pi(a|s)A^\pi(s, a)\big]$ \cite{mniha16a3c, schulman2016} with respect to the parameters of $\pi$, where $d^{\pi}(s)$ is the marginal state density induced by $\pi$, and the advantage function $A^\pi$ can be estimated as $\mathcal{T}^\pi V^\pi(s) - V^\pi(s)$.

\subsubsection{Distributional RL}
In distributional reinforcement learning, the distributions of returns instead of their expectations (i.e. value functions) are maintained. The distributional Bellman equation in terms of the state-action return is \cite{bellemare17}
\begin{equation}\label{eq:bellmanop_drl_q}
\mathcal{T}^\pi G^\pi(s, a):\stackrel{D}{=}R(s, a)+\gamma G^\pi(S', A').
\end{equation}
The distribution equation $U:\stackrel{D}{=}V$ specifies that the random variable $U$ is distributed by the same law as is random variable $V$. The reward $R(s, a)$, next state-action tuple $S', A'$ and its return $G^\pi(S', A')$ are random variables, with compound randomness stemmed from $\pi, P$, and $\mathscr{R}$.

Eq. \eqref{eq:bellmanop_drl_q} is a contraction mapping in the $p$-th order Wasserstein metrics $W_p$ \cite{bellemare17}.
Previously, Eq. \eqref{eq:bellmanop_drl_q} is exploited in a $Q$-learning style value iteration,
with the distribution of $G^\pi(s, a)$ represented as a particle set, updated either through cross-entropy loss \cite{bellemare17},
quantile regression \cite{dabney18, dabney18QR, rowland19},
or Sinkhorn iterations \cite{martin20}.
Particle-based (ensemble-)critics $G^\pi(s, a)$ are incorporated into conventional off-policy policy gradient methods
by \cite{barthmaron18} and \cite{kuznetsov20}. A continuous $G^\pi(s, a)$ distribution can be conferred via Wasserstein-GAN (WGAN) \cite{arjovsky17}, and has been investigated
in both $Q$-learning \cite{doan18} and policy gradients \cite{freirich19}. We remark that these works all estimate return distributions with empirical approximations, e.g. particle set or WGAN.

\section{Methods}
We begin with proving that the distributional Bellman operation in terms of \textit{state}-return distributions is also a contraction mapping in Wasserstein metrics.
We then show resemblance between distributional Bellman update and a variational Bayesian solution to return distributions, leading to a novel distributional RL approach. Thereafter, we propose an internal incentive that leverages posterior IG stemmed from return estimation inaccuracy.

\subsection{Distributional Bellman Operator for \textit{State}-Return}
First, in the same sense that Eq. \eqref{eq:bellmanop_drl_q} extends Eq. \eqref{eq:bellmanop_convention_q}, we extend Eq. \eqref{eq:bellmanop_convention} and define the distributional Bellman operator regarding the state return $G^\pi(s)$ as
\begin{equation}\label{eq:bellmanop_drl}
\mathcal{T}^\pi G^\pi(s):\stackrel{D}{=}R(s)+\gamma G^\pi(S').
\end{equation}
Now we demonstrate that Eq. \eqref{eq:bellmanop_drl} is also a contraction in $p$-Wasserstein metrics.

For notional convenience, we write the infimum-$p$-Wasserstein metric in Eq. \eqref{eq:wass_inf} in terms of random variables: $d_p(X, Y) := W_p(\alpha, \beta), X\sim\alpha, Y\sim\beta$.

Let $\mathcal{G}\in\mathbb{R}^n$ denote a space of returns valid in the MDP, and $\Omega\in\mathscr{P}(\mathcal{G})(\mathcal{S})$ a space of state-return distributions with bounded moments. Represent as $\omega$ the collection of distributions $\{\omega(s)\big|s\in\mathcal{S}\}$, in which $\omega(s)$ is the distribution of random return $G(s)$. For any two distributions $\omega_1, \omega_2\in\Omega$, the supremum-$p$-Wasserstein metric on $\Omega$ is defined as \cite{bellemare17, rowland18}
\begin{equation}
\bar{d}_p (\omega_1, \omega_2) := \mathop{\mathrm{sup}}_{s\in\mathcal{S}} d_p\big(G_1(s), G_2(s)\big).
\end{equation}

\noindent
\textbf{Lemma 2.} $\bar{d}_p$ \textit{is a metric over state-return distributions.}\\

The proof is a straightforward analogue to that of Lemma 2 in \cite{bellemare17}, substituting the state space $\mathcal{S}$ for state-action space $\mathcal{S}\times\mathcal{A}$.\\ 

\noindent
\textbf{Proposition 1.} \textit{The distributional Bellman operator for state-return distributions is a} $\gamma$\textit{-contraction in} $\bar{d}_p$.
\begin{proof}
The reward $R(s)\in\mathcal{G}$ is a random vector such that $R(s) = \int_{\mathcal{A}} \mathscr{R}(s, a)\tilde{\pi}(a|s)\mathrm{d}a$, where $\tilde{\pi}$ denotes the normalised policy $\pi$.

\noindent
Represent the marginal state transition kernel under policy $\pi$ as $P^\pi(s'|s) = \int_{\mathcal{A}} P(s'|s, a)\tilde{\pi}(a|s)\mathrm{d}a$. Then define a corresponding transition operator $\mathcal{P}^\pi: \mathcal{G}\mapsto\mathcal{G}$
\begin{equation}
\mathcal{P}^\pi G(s) :\stackrel{D}{=} G(S'), ~S'\sim P^\pi(\cdot|s).
\end{equation}

\noindent
With the marginal state transition operator substituted for the action-dependent one, the rest of the proof is analogous to that of Lemma 3 presented by \cite{bellemare17}. 
\end{proof}

We therefore conclude that Eq. \eqref{eq:bellmanop_drl} has a unique fixed point $G^\pi$. Proposition 1 vindicates backing up distributions of the state return $G^\pi(s)$ by minimising Wasserstein metrics to a target distribution.

For the policy gradient theorem 
\cite{62Sutton1999PGM} to hold, one would need at each encountered $s_t$ an unbiased estimator of $\mathbb{E}{\big[\sum_{k=0} \gamma^k r_{t+k}\big]}$ in computing the policy gradient. In distributional RL, such a quantity is obtained by sampling from the approximated return distribution (or averaging across such samples). The Bellman operator being a contraction ensures convergence to a unique true on-policy return distribution, whose expectation is thereby also unbiased. The same holds also for sample estimates of the state-conditioned reward-to-go and thereby for the advantage function.

\begin{algorithm}[t]
\caption{Bayesian Distributional Policy Gradients} \label{alg:bd}
\begin{algorithmic}[1]
\State Initialise encoder $q_\phi(Z|X, S)$, generator $G_\theta(Z, S)$, prior $p_\theta(Z|S)$, discriminator $D_\psi(X, Z, S)$ and policy $\pi$

\State \textbf{While} not converge:

\footnotesize\texttt{// roll out}

\State $~~~~~~$ training batch $\mathcal{D}\gets\varnothing$

\State $~~~~~~$ \textbf{For} $t=0, \dots, k-1, ~\forall$ threads:

\State $~~~~~~~~~~~~$ execute $a_t\sim\pi(\cdot|s_t)$, get $r_t$, $s_{t+1}$

\State $~~~~~~~~~~~~$ sample return $z_t\sim p_\theta(\cdot|s_t), ~g_t\gets G_\theta(z_t, s_t)$

\State $~~~~~~~~~~~~$ update $\mathcal{D}$

\State $~~~~~~$ sample last return $z_k\sim p_\theta(\cdot|s_k), ~g_k\gets G_\theta(z_k, s_k)$

\footnotesize\texttt{// estimate advantage for whole batch}

\State $~~~~~~$ \textbf{For} $t \in \mathcal{D}$:

\State $~~~~~~~~~~~~$ estimate advantage $\hat{A}_t$ with $r_{t:t+k-1}, ~g_{t:t+k}$

$~~~~~~$ using any estimation method

\State $~~~~~~~~~~~~$ Bellman target $x_t\gets \hat{A}_t + g_t$

\State $~~~~~~~~~~~~$ get curiosity reward $r^c_t$ by Eq.\eqref{eq:z_info_gain}-\eqref{eq:gross_uncty_end}

\State $~~~~~~~~~~~~$ get augmented advantage $\hat{A}_t^c$ by substituting $r_t$ in $\hat{A}_t$

$~~~~~~$ with $r_t + r^c_t$

\footnotesize\texttt{// train with mini batch $B\subset\mathcal{D}$}

\State $~~~~~~$ \textbf{For} $t \in B$:

\State $~~~~~~~~~~~~$ encode $\tilde{z}_t\sim q_\phi(\cdot|x_t, s_t)$

\State $~~~~~~~~~~~~$ sample from joint $p_\theta$: $z_t\sim p_\theta(\cdot|s_t), \tilde{x}_t\gets G_{\bar{\theta}}(z_t, s_t)$

\footnotesize\texttt{// take gradients}

\State $~~~~~~$ update $D_\psi$ by ascending

$\frac{1}{|B|} \sum_{t\in B} \log D_\psi(\tilde{x}_t, z_t, s_t)
+ \log \big(1 - D_\psi(x_t, \tilde{z}_t, s_t)\big)$

\State $~~~~~~$ update encoder, prior by ascending

$\frac{1}{|B|} \sum_{t\in B} \log \big(1 - D_\psi(\tilde{x}_t, z_t, s_t)\big)
+ \log D_\psi(x_t, \tilde{z}_t, s_t)$

\State $~~~~~~$ update $G_\theta$ by descending $\frac{1}{|B|} \sum_{t\in B} ||x_t-G_\theta(\tilde{z}_t, s_t)||^2_2$

\State $~~~~~~$ update $\pi$ by ascending $\frac{1}{|B|} \sum_{t\in B} \log \pi(a_t|s_t)\hat{A}^c_t$

using any policy gradient method

\State Return $\pi$
\end{algorithmic}
\end{algorithm}

\subsection{Inference in Distributional Bellman Update}
We now proceed to show that the distribution of $\mathcal{T}^\pi G^\pi(s)$ can be interpreted as the target distribution $p_X$, and hence propose a new approach to distributional RL. Specifically, let the data space $\mathcal{X}=\mathcal{G}$ be the space of returns. $\forall ~s\in\mathcal{S}$, we shorthand as such
$x(s):= \mathcal{T}^\pi G^\pi(s), ~g(s):= G^\pi(s)$,
thus $x(s), g(s)\in \mathcal{X}$. We view the Bellman target $x(s)$ as a sample from the empirical data distribution $x(s)\sim p_X(X|s)$, whilst the estimated return $g(s)$ is generated from the model distribution $g(s)\sim p_G(X|s)$. The state $s$ is an observable condition to the generative model: its distribution is of no interest to and not modelled in the Bayesian system.

We factorise the $s$-conditioned sampling distributions in Lemma 1 such that
\begin{align}\label{eq:factoriz}
p_\theta(X, Z|s) &:= p_\theta(X|Z, s)p_\theta(Z|s), \nonumber\\
q_\phi(X, Z|s) &:= p_X(X|s)q_\phi(Z|X, s).
\end{align}
$p_\theta(X|Z, s)=\delta_{G_\theta(Z, s)}(X)$ is a deterministic decoder.

The intuition of a state-conditioned, learned prior for $Z$ instead of a simple, fixed one, is to add stochasticity for the prior and encoder to meet halfway.
Similar to the encoder, we represent the prior also in a variational fashion and sample through re-parameterisation during gradient estimation.

Lemma 1 implies that training $D_\psi$ and the generative model alternatingly with factorisation in Eq. \eqref{eq:factoriz} would suffice to both have the encoder $q_\phi(Z|X, s)$ approximating the true posterior $p_\theta(Z|X, s):= p_\theta(X|Z, s) p_\theta(Z|s)/p_X(X|s)$ and to reconstruct in $\mathcal{X}$ \cite{dumoulin17ali, donahue17bigan}. Notice that a globally observable condition $s$ is orthogonal to Lemma 1 and Eq. \eqref{eq:wae}. And so is a learned prior: both the true posterior and the push-forward are \textit{relative to} the prior $p_\theta(Z|s)$.

In our work, in contrast, $G_\theta$ is deemed fixed in relation to the minimax game,
leaving the \textit{encoder}, \textit{prior} and \textit{discriminator} to be trained in the minimax game
\begin{align}\label{eq:gan_method}
\mathop{\mathrm{max}}_{D_\psi} &\mathop{\mathrm{min}}_{q_\phi, p_\theta} ~\mathbb{E}_{z\sim p_\theta(Z|s)} \big[\log D_\psi(G_{\bar{\theta}}(z, s), z, s)\big] \nonumber\\
&+ \mathbb{E}_{x\sim p_X(X|s)} \mathbb{E}_{z\sim q_\phi(Z|x, s)} \big[\log \big(1-D_\psi(x, z, s)\big)\big].
\end{align}
The overhead bar $\bar{(\cdot)}$ denotes that gradient is not backpropagated through the parameter in question. This means $q_\phi(Z|X, s)$ is still trained to approximate the true posterior induced by the current $G_\theta$, irrespective of the capability of the latter for reconstruction. Meanwhile, the reconstruction is achieved by minimising a Wasserstein metric in $\mathcal{X}$
\begin{equation}\label{eq:wae_method}
\mathop{\mathrm{min}}_{G_\theta} \mathbb{E}_{x\sim p_X(X|s)}\mathbb{E}_{z\sim q_{\bar{\phi}}(Z|x, s)} \big[d^p\big(x, G_\theta(z, s)\big)\big].
\end{equation}

Essentially, we are alternating between training the encoder and prior via Eq. \eqref{eq:gan_method} and training the generator via Eq. \eqref{eq:wae_method}.
We will use a fixed prior $p_Z$ and omit state dependence in the ensuing discussion, as they do not affect convergence.
If the encoder approximates the true posterior everywhere in $\mathcal{X}$, the aggregated posterior $Q(Z)$ is naturally matched to the prior $p_Z(Z)$, so long as $p_\theta(X|Z)$ is properly normalised, as is indeed the case when it's degenerate. As such, meeting the constraint on the search space in Eq. \eqref{eq:wae} is a necessary condition to accurate posterior approximation.

\begin{figure*}[tbh]
    \centering
    \includegraphics[scale = 0.38]{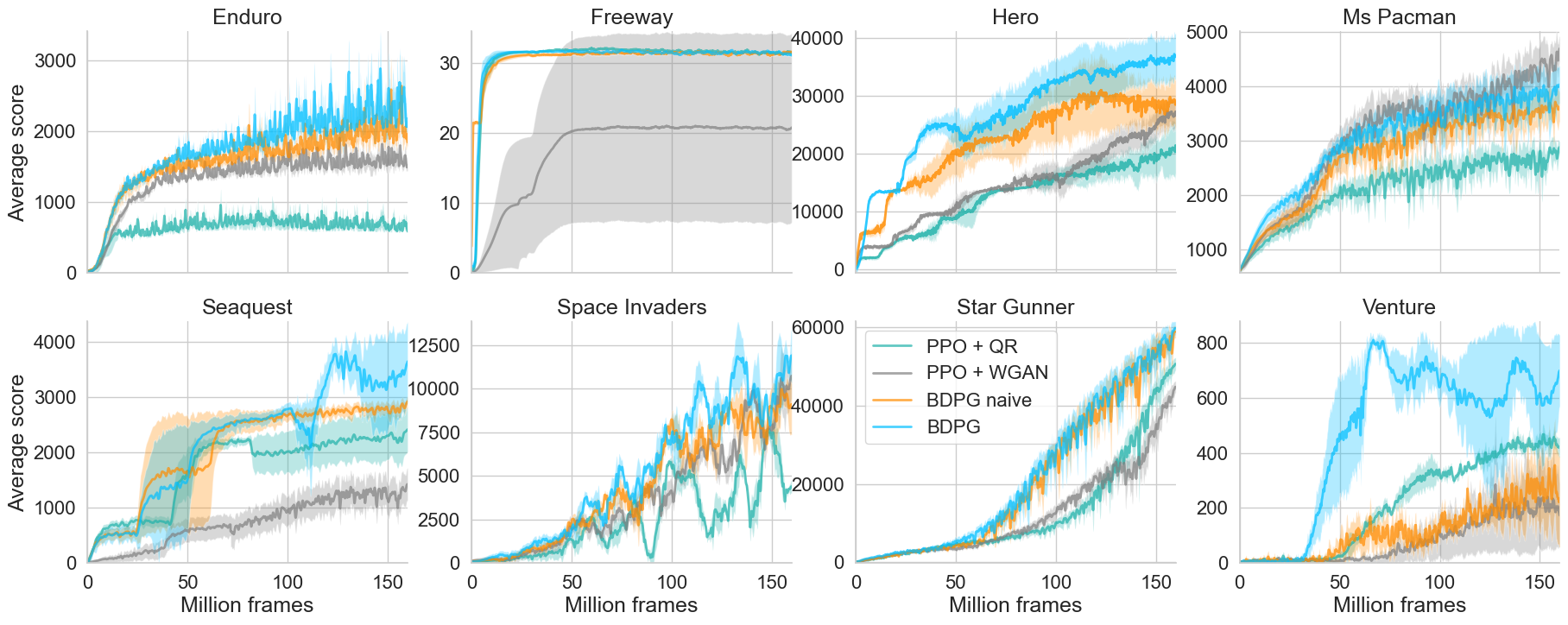}
    \caption{Learning curves on Atari games with the mean (solid line) and standard deviation (shaded area) across $5$ runs.}
    \label{fig:atari}
\end{figure*}

Note that in Eq. \eqref{eq:wae}, $\mathbb{E}_{p_X}\mathbb{E}_{q_\phi}[ G_\theta(Z)]$ is the push-forward of $Q(Z)$, $G_{\theta\#}Q$. The primal form of $W_p\big(p_X, p_G\big)$, where $p_G = G_{\theta\#}p_Z$, is thereby the infimum of $W_p\big(p_X, G_{\theta\#}Q\big)$ over $q_\phi$ s.t. $Q=p_Z$. Therefore, $W_p(p_X, G_{\theta\#}Q)$ is an upper bound to the true objective $W_p(p_X, G_{\theta\#}p_Z)$ upon $Q=p_Z$.

Learning converges as we explain in the following. And to provide intuition, we highlight the resemblances to the Expectation-Maximisation (EM) algorithm.
Eq. \eqref{eq:gan_method} enforces contrastive learning such that the variational posterior approaches the true posterior, comparable to the E-step in EM. Eq. \eqref{eq:gan_method} allows to compute a bound $W_p(p_X, G_{\theta\#}Q)$ in Eq. \eqref{eq:wae_method}, which is equivalent to the computationally tractable surrogate objective function of the negative free energy in EM, or ELBO in variational Bayes. The expected Wasserstein metric w.r.t. the current $q_\phi$ is then minimised by updating the parameters of the decoder via Eq.~\eqref{eq:wae_method}. This update is reminiscent of the M-step in  EM, which maximises the expected log likelihood while fixing inference for $Z$.

In our method  $W_p(p_X, G_{\theta\#}Q)$ acts as an upper bound when $Q=p_Z$, whereas in EM the surrogate objective is a lower bound. This upper bound decreases in Eq. \eqref{eq:gan_method} as it approaches the true objective $W_p(p_X, G_{\theta\#}p_Z)$. Eq. \eqref{eq:wae_method} then decreases $W_p(p_X, G_{\theta\#}Q)$ further and consequently also decreases $W_p(p_X, G_{\theta\#}p_Z)$. Note, the condition $Q=p_Z$ does not have to hold on each iteration, but can be amortised over iterations.
Assuming infinite model expressiveness, the discrepancy between $Q$ and $p_Z$ shrinks monotonically, as all determinant functions for $Q:=\mathbb{E}_{p_X}[q_\phi(\cdot|x)]=p_Z$ in Eq. \eqref{eq:gan_method} are fixed irrespective of the value of $\theta$. When $q_\phi$ converges to the true posterior, $W_p(p_X, G_{\theta\#}Q)$ is more sufficiently an upper bound due to restricted search space in the primal form. 
While $W_p(p_X, G_{\theta\#}Q)$ functions as an amortised upper bound, $W_p(p_X, G_{\theta\#}p_Z)$ still decreases continually (as opposed to from each iteration) and converges to a local minimum.

The merit of the two-step training is two-fold: 1) with only the distributions over $Z$ under tuning in the minimax game, the adversarial training comes off with a weaker topology and is not relied upon for reconstruction, making its potential instability less of a concern;
and 2) an explicit distance loss $d$ in $\mathcal{X}$ minimises $W_p$ to ensure contraction of return distribution backups.
If everything was trained adversarially in JS divergence and allowed to reach global optimum, 
the decoder and encoder would be reversing each other both in density domain. In our setting, 
$Q$ is matched to $p_Z$ everywhere in $\mathcal{Z}$, while $p_G$ has minimum $W_p$ distance to $p_X$.

At each step of environmental interaction, a state return is sampled via the standard two steps $g(s)\sim p_G(X|s) \Longleftrightarrow z\sim p_\theta(Z|s), g(s)\gets G_\theta(z, s)$. The one-step Bellman target $x(s)$ is calculated as $r + \gamma g(s')$. Generalisation to $k$-step bootstrap can be made analogously to the conventional RL.

\subsection{Exploration through Posterior Information Gain}
Curiosity \cite{schmidhuber1991, schaul16, houthooft16, freirich19} produces internal incentives when external reward is sparse.
We explore through encouraging visits to regions where the model's ability to predict the reward-to-go from current return distribution is weak. However, the Bellman error $x(s)-g(s)$ is not a preferable indicator, as high $x(s)-g(s)$ may well be attributed to high moments of $g(s)$ itself under point estimation (i.e. the \textit{aleatoric} uncertainty), whereas it is the uncertainty in value belief due to estimating parameters with limited data around the state-action tuple (i.e. the \textit{epistemic} uncertainty) that should be driving strengthened visitation.

To measure the true reduction in uncertainty about return prediction, we estimate discrepancies in function space instead of parameter space. Specifically, the insufficiency in data collection can be interpreted as how much a posterior distribution of a statistic or parameter inferred from a condition progresses from a prior distribution with respect to the action execution that changes this condition, i.e., the IG. A large IG means a large amount of data is required to achieve the update. In its simplest form, the condition is implicitly the data trained on. In exact Bayes, the condition itself can be thought of as a variable estimated from data, e.g. the random return $X$, hence enabling an explicit IG derived from existing posterior model $q_\phi(Z|X)$. Therefore, we define the IG $u(s)$ at $s$ in return estimation as
\begin{equation}\label{eq:z_info_gain}
u(s) := \mathrm{KL}\big(q_\phi\big(Z|x(s),s\big)\big|\big|q_\phi\big(Z|g(s),s\big)\big).
\end{equation}

Before the transition, the agent's estimation for return is $g(s)$. The action execution enables the computation of the Bellman target $x(s)$, which would not be viable before the transition, in which $q_\phi\big(Z|g(s),s\big)$ acts here as a prior. As a result, $u(s)$ would encourage the agent to make transitions that maximally acquire new information about $Z$, hence facilitating updating $p_G$ towards $p_X$.
Upon convergence, $g(s)$ and $x(s)$ are indistinguishable and the IG approaches $0$. The benefit of our IG is tree-fold: it is moments-invariant, makes use of all training data, and increases computation complexity only in forward-passing the posterior model when calculating the KL divergence without even requiring gradient backpropagation.

The curiosity reward $r^c(s)$ is determined by $u(s)$ and a truncation scheme $\mathcal{R}:\mathbb{R}^+\mapsto\mathbb{R}^{[0, ~\eta\cdot \bar{u})}, ~\eta, \bar{u}\in\mathbb{R}^+_*$, to prevent radical exploration
\begin{equation}\label{eq:gross_uncty_end}
r^c(s) := \mathcal{R}(u(s)) := \eta_t\cdot\mathrm{min}~\big(u(s), ~\bar{u}\big).
\end{equation}

We exploit relative value by normalising the clipped $u(s)$ by a running mean and standard deviation of previous IGs. The exploration coefficient $\eta_t$ is logarithmically decayed as $\eta_t=\eta\sqrt{\log t / t}$, by the rate at which the parametric uncertainty decays \cite{koenker05QR, mavrin19}, where $t$ is the global training step, and $\eta$ an initial value, to assuage exploration getting more sensitive to the value of $u(s)$ as parameters become more accurate.

We use $r^c$ to augment return backup during policy update, as the purpose is for the action to lead to uncertain regions by encouraging curiosity about future steps. When training the generative model for return distributions we use the original reward only.

We investigate a multi-step advantage function. The contraction property of the distributional Bellman operator is propagated from $1$-step to $k$-step scenarios by the same logic as in conventional RL.
The benefit of looking into further steps for exploration is intuitive viewed from the long-term goal of RL tasks: the agent should not be complacent about a state just because it is informative to immediate steps.


The pseudocode in Algorithm \ref{alg:bd} presents a mini-batch version of our methodology BDPG. We denote state return $g(s_t)$ as $g_t$ for compactness. Other step-dependent values are shorthanded accordingly. We use Euclidean distance for reconstruction, leading to the $W_2$ metric being minimised. $k$ is the number of unroll steps, and is also the maximum bootstrap length, albeit the two are not necessarily the same. 

\section{Related Work}
Policy optimisation enables importance sampling based off-policy evaluation for re-sampling weights in experience replay schemes \cite{Wang16sampleeff, gruslys18}. In continuous control, where the policy is usually a parametric Gaussian, exploration can be realised by perturbing the Gaussian mean \cite{lillicrap15ddpg, ciosek19}, or maintaining a mixture of Gaussians \cite{lim18actorexpert}.
Alternatively, random actions can be directly incentivised by regularising policy (cross-)entropy \cite{abdolmaleki15, mniha16a3c, nachum16, akrour16, haarnoja18sac}.

\begin{figure}[th]
    \centering
    \includegraphics[scale = 0.37]{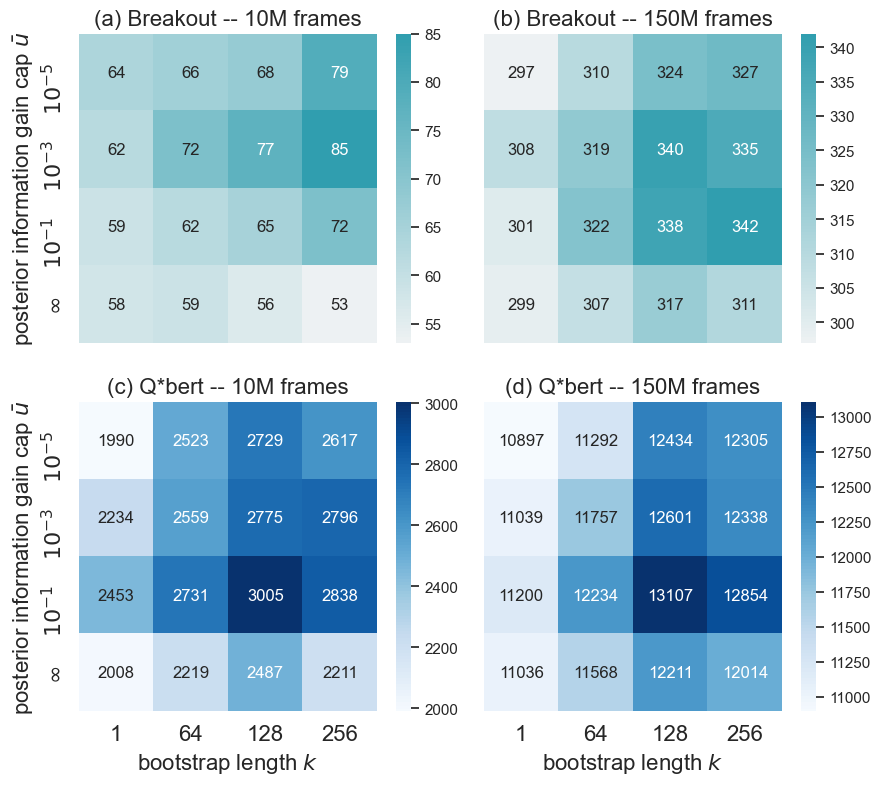}
    \caption{Impact of bootstrap length $k$ and truncation cap $\bar{u}$ for information gain
    at $10$M and $150$M steps into training.
    }
    \label{fig:abl}
\end{figure}

A group of works propose to exploit epistemic uncertainty via an approximate posterior distribution of $Q$ values. Stochastic posterior estimates are constructed through training on bootstrapped perturbations of data \cite{osband16bootstrappeddqn, osband19}, or overlaying learned posterior variance \cite{chen17ucb, odonoghue18}. While this series of works can be thought of as posterior sampling w.r.t. $Q$ values, \cite{tang18} approximates Bayesian inference by sampling parameters for a distributional RL model. On the other hand, a particle-based distributional RL model itself registers notion of dispersion, inspiring risk-averse and risk-seeking policies \cite{dabney18} and optimism-in-the-face-of-uncertainty quantified by the variance of the better half of the particle set \cite{mavrin19}.

There are also approaches exploiting dynamics uncertainty \cite{houthooft16}, pseudo counts \cite{bellemare16count, ostrovski17, tang17}, gradient of a generative model \cite{freirich19}, and good past experiences \cite{oh2018SIL}, that do not estimate dispersion or model disagreement of value functions.

\begin{figure*}[tbh]
    \centering
    \includegraphics[scale =0.35]{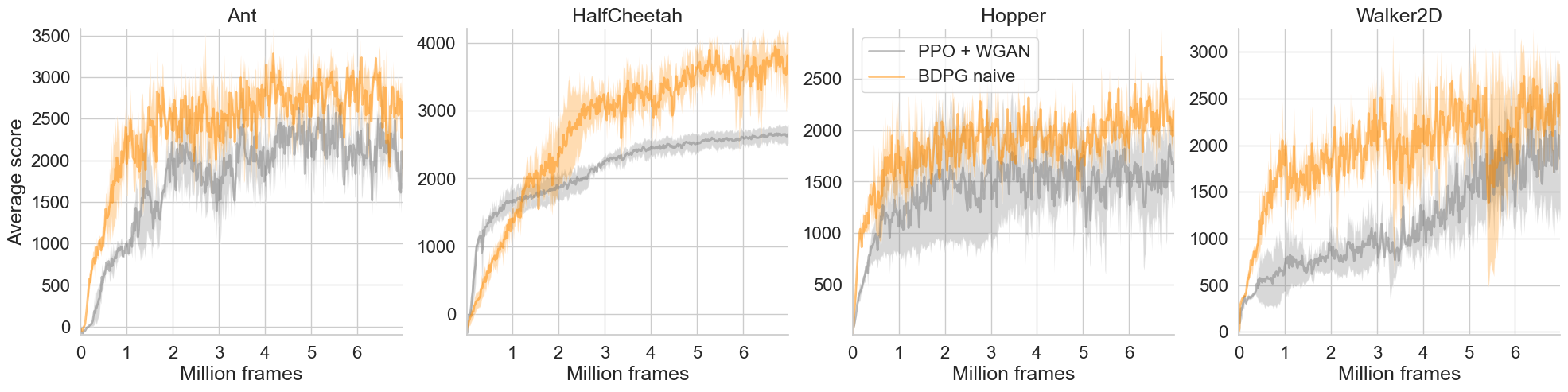}
    \caption{Learning curves on MuJoCo tasks with the mean (solid line) and standard deviation (shaded area) across $5$ runs.}
    \label{fig:mjc}
\end{figure*}

\section{Evaluation}
We evaluate our method on the Arcade Learning Environment Atari 2600 games and continuous control with MuJoCo. We estimate a $k$-step advantage function using Generalised Advantage Estimation (GAE) \cite{schulman2016}, and update the policy using Proximal Policy Optimisation (PPO) \cite{schulman17ppo} which maximises a clipped surrogate of the policy gradient objective. 
For both Atari and MuJoCo environments, we use $16$ parallel workers for data collection, and train in mini batches. For Atari, we unroll $128$ steps with each worker in each training batch for all algorithms, and average scores every $80$ training batches. For MuJoCo, we unroll $256$ steps, and average scores every $4$ batches. Except for ablation experiments that used rollout length $\mathrm{max}(k, 128)$, the number of unroll steps is also the bootstrap length $k$.

We compare to other distributional RL baselines on eight of the Atari games, including some of the recognised hard-exploration tasks: $\mathrm{Freeway}$, $\mathrm{Hero}$, $\mathrm{Seaquest}$ and $\mathrm{Venture}$.

Direct comparisons to previous works are not meaningful due to compounded discrepancies. To allow for a meaningful comparison, we implement our own versions of baselines, fixing other algorithmic implementation choices such that the tested algorithms vary only in how return distributions are estimated and in the exploration scheme. We modify two previous algorithms \cite{freirich19} and \cite{dabney18QR}, retaining their return distribution estimators as benchmarks: a generative model Wasserstein-GAN \cite{arjovsky17} (PPO+WGAN), and a discrete approximation of distribution updated through quantile regression \cite{koenker05QR} (PPO+QR). Importantly, all of our baselines are \textit{distributional} RL solutions that maintain \textit{state}-return distributions.

Our BDPG is evaluated in two versions: 1) with the naive add-noise-and-argmax \cite{mniha16a3c} exploration mechanism (BDPG naive), and 2) one that explores with the proposed curiosity reward (BDPG). Naive exploration is also used for the PPO+WGAN and PPO+QR baselines. Learning curves in Figure \ref{fig:atari} suggest that with exploration mechanism fixed, the proposed Bayesian approach BDPG naive outperforms or is comparable to WGAN and QR in $6$ out of $8$ games. Morever, BDPG is always better or equal to BDPG naive, vindicating our exploration scheme, and is able to get the highest score among all tested algorithms in all four of the hard-exploration games tested on.

We conduct ablation and parameter studies to investigate the impact of the bootstrap length $k$, and of the truncation cap $\bar{u}$ on the IG $u(s)$, on Atari games $\mathrm{Breakout}$ and $\mathrm{Q}^*\mathrm{bert}$. In particular, $k=1$ and $\bar{u} \rightarrow \infty$ are looked at as ablation cases. Average scores of the batch started at $10$M and $150$M steps into training are shown in Figure \ref{fig:abl}. Each coloured pixel corresponds to the best outcome with respect to $\eta$ value among its selection sweep according to average long-term performance for each combination of $k$ and $\bar{u}$. We found that as training progresses, short $k$ comes to display a more prohibitive effect, as the model becomes more discriminative about the environment, and lack of learning signals, i.e. fewer rewards to calculate the Bellman target with, becomes increasingly suppressive.
Our experiments suggest that although the best bootstrap length depends on the task, longer bootstrapping generally produces better long-term performance. But a long bootstrap length does not work well with a large $\bar{u}$, a possible explanation is that as $k$ increases, the variance in the Bellman targets multiplies. In this scenario, the agent may encounter states with which it is very unfamiliar.
The value of $u(s)$ can grow unbounded and the tendency to explore get out of hand if we do not curb it.
Moreover, such extreme values can also jeopardise subsequent curiosity comprehension through the normalisation of $u(s)$. This phenomenon justifies the application of our truncation scheme, especially for larger $k$. In addition, we found that choosing too large $k$ does not diminish performance drastically, potentially due to the return distribution already accounting for some degree of reward uncertainty, which is a helpful characteristic when prototyping agents.

In the continuous control tasks with MuJoCo,
we focus on the ability of distributional RL algorithms to generalise, and less the challenge of exploration.
Therefore, we compare the performance of BDPG naive against the benchmark distributional RL algorithm PPO+WGAN, a generative solution that does not conduct inference. Both are stripped of exploration incentives. Noticeable amounts of variance displayed throughout training for both algorithms may be due to that they both involve adversarial training. As shown in Figure \ref{fig:mjc}, however, our model outperforms the benchmark in all cases with distinct margins. We believe this is because WGAN does not take expectations across an amortised inference space that accounts for better generalisation. This proves to be highly beneficial for reasoning about return distributions in continuous control tasks such as MuJoCo environments, where robustness in the face of unseen data weighs up more in behaviour stability.

\section{Conclusion}
We formulate the distributional Bellman operation as an inference-based 
auto-encoding process. We demonstrated contraction of the Bellman 
operator for state-return distributions, expanding on the distributional
RL body of work that focused on state-action returns, to date. 
Our tractable
solution alternates 
between minimising Wasserstein metrics to continuous distributions of the 
Bellman target and adversarial training for joint-contrastive learning to
estimate a variational posterior from bootstrapped targets.
This allows us to distill the benefits of Bayesian inference into distributional RL, where predictions of returns are based on expectation and thus more accurate
in the face of unseen data.
As a second innovation we use the availability of a variational posterior to derive a curiosity-driven exploration mechanism, which we show is more efficiently solving hard-exploration tasks. Either of our two contributions can be combined with other building blocks to form new RL algorithms, e.g. in Explainable RL \cite{BeyretSF19}. We believe that our innovations link and expand the applicability and efficiency of distributional RL methods.

\section{Acknowledgments}
We are grateful for our funding support: a Department of Computing PhD Award to LL
and a UKRI Turing AI Fellowship (EP/V025449/1) to AAF.

\end{document}